\documentclass[a4paper]{jpconf}
\usepackage{graphicx}

\newcommand{\dee}[0]{\, \mathrm{d}}
\newcommand{\deepow}[1]{\, \mathrm{d}^#1}
\usepackage{amsfonts}
\usepackage{amsmath}

\begin{document}
\title{Green's function based unparameterised multi-dimensional kernel density and likelihood ratio estimator}

\author{P K\"oves\'arki, I C Brock and A E Nuncio Quiroz}
\address{Physics Institute, University of Bonn}
\ead{kovesarki@physik.uni-bonn.de, brock@physik.uni-bonn.de, nuncio@physik.uni-bonn.de}

\begin{abstract}
This paper introduces a probability density estimator based on Green's function identities. A density model is constructed under the sole assumption that the probability density is differentiable. The method is implemented as a binary likelihood estimator for classification purposes, so issues such as mis-modeling and overtraining are also discussed. The identity behind the density estimator can be interpreted as a real-valued, non-scalar kernel method which is able to reconstruct differentiable density functions. 
\end{abstract}

\section{Theory of density estimation}

\label{sec:Theory}

Generally, a density estimator is an identity operator for the probability density function. If it is a linear operator, then it is an approximation of the Dirac-delta function:

$$g_{\text{est}}(x) = \int \delta_{\text{approx}}(x-x') g(x') \dee x\,.$$ 
Fixed kernel methods usually estimate the density with a bias, since the kernel does not converge to the delta function. The same can be seen on histograms with fixed bin sizes, and to make them unbiased certain adaptivity is necessary. 
Different kernels usually maintain different features of the $g_\text{est}$, such as differentiability or easy recognizability of peaks~\cite{Bishop:2006:PRM:1162264}.
Non-adaptive methods are the estimation from the moments via Fourier integrals~\cite{James:2006zz} or the estimations that target the integral probability density.
A problem with the methods using Fourier integrals, is that they may require computationally expensive volume integrals, and the result is not necessarily a density function until sufficient number of moments are known. 

To avoid the complexity of finding the global minimum of the error function, most of the overtraining problems and heavy volume integrals, we tried to define a successive adaptive kernel method to estimate the probability densities, using a kernel that is suppressed at large distances. The starting point is, that functionals like linear operators are preferred, as they can be successfully approximated with a sample of a distribution, because they act like expectation values of random variables.
$$ \lim_{N\rightarrow \infty} \sum_{i=1}^{N} Lp = \int L(x,x') p(x')  \dee \mu(x')\,,$$ 
where $L$ is a linear operator, $p$ is a parameter vector and $\mu(x)$ is a probability measure. To have such an estimator for the probability density itself, an identity operator is needed:

$$ \mu(x) =I(\mu) = P\left( \int L(x,x') p(x')  \dee \mu(x') \right)\,,$$
where $P$ is a necessary post-processing to convert the linear operator's output to a scalar. Identity operators can be constructed using Green functions of operators. A particularly interesting one is the Green function of the Laplace operator in $n\geq 3$ dimensions\footnote{Although the Green's function is different for $n=2$, the same procedure can be repeated and it gives the same form of functions at the end, making method valid in 2 dimensions.}. 

\begin{equation} \label{eq:LaplaceGreen}
 \Delta G(x,x') = \delta(x-x')
 \end{equation}
$$  G(x,x') = -\frac{1}{S_n}\frac{1}{|x-x'|^{n-2}}\,.$$
Here $S_n$ is a constant, the surface of a unit sphere in $n$ dimensions. For twice differentiable functions $c(x)$ that disappear at the boundaries, $\lim \limits_{x\rightarrow \infty} c(x) = 0$, the identity operator can be constructed:

\begin{align*}
c(x)  = & -\frac{1}{S_n}\int \limits_{\mathbb{R}^n} \frac{\Delta c(x')}{|x-x'|^{n-2}} \deepow{n} x'\\
	 = & \frac{1}{S_n}\int \limits_{\mathbb{R}^n} \partial_{\mu'} \frac{1}{|x-x'|^{n-2}} \cdot \partial_{\mu'} c(x') \deepow{n} x'   \,.
\end{align*} 
This can not be used directly for density estimations, since it is linear in $\partial_{\mu'}c(x')$ instead of $c(x')$, but an identity can be constructed for the vector field of the first derivative:

$$ \partial_{\mu} c(x) = \frac{1}{S_n}\int \limits_{\mathbb{R}^n} \partial_{\mu} \partial_{\mu'} \frac{1}{|x-x'|^{n-2}} \cdot \partial_{\mu'} c(x') \deepow{n} x' \,.$$
The amplitude of $\partial_{\mu}c(x)$ is still a scalar function; hence for those probability measures, $g(x)$, where a $c(x)$ exists\footnote{See the subsection targeted at the existence of such a $c(x)$.} with $|\partial_\mu c(x)| = g(x)$, there must be at least one $\phi_\mu(x)$ vector field consisting of $\phi_\mu\phi^\mu = 1 $ unit vectors, for which the following identity holds:

\begin{equation} \label{eq:identityfordensity}
g(x) \phi_\mu(x) = \frac{1}{S_n}\int \limits_{\mathbb{R}^n} \partial_{\mu} \partial_{\mu'} \frac{1}{|x-x'|^{n-2}} \cdot \phi_{\mu'}(x')  g(x')  \deepow{n} x'   \,.
\end{equation}

Since the integral operator was derived from the Laplace operator, in three dimensions it is similar to the \emph{dipole} interaction operator in electromagnetism, hence $\phi_\mu$ might be called a dipole field. 
What is important is that this $\phi_\mu(x)$ field can be found without the knowledge of the true $g(x)$; only the possibility of calculating the expectation values on the left-hand side of eq.~\eqref{eq:identityfordensity} is required. In this case it is possible to calculate the left-hand side for any configuration of $\phi_{\mu'}(x')$, until the expectation value $E_\mu(x)$ has the same direction as $\phi_\mu(x)$ at every $x$:

\begin{equation}\label{eq:StoppingCriterium}
\phi_\mu(x) \parallel E_\mu(x) = \frac{1}{S_n}\int \limits_{\mathbb{R}^n} \partial_{\mu} \partial_{\mu'} \frac{1}{|x-x'|^{n-2}} \cdot \phi_{\mu'}(x')  g(x')  \deepow{n} x'  \,.
\end{equation}
A way to show that $E_\mu(x) \approx g(x)\phi_\mu(x)$ is to do an integration by parts first:

$$ E_\mu(x) = \frac{1}{S_n}\int \limits_{\mathbb{R}^n} \partial_{\mu} \partial_{\mu'} \frac{1}{|x-x'|^{n-2}} \cdot \phi_{\mu'}(x')  g(x')  \deepow{n} x' =  
	-\frac{1}{S_n}\int \limits_{\mathbb{R}^n} \partial_{\mu} \frac{1}{|x-x'|^{n-2}} \cdot \partial_{\mu'} \left(\phi_{\mu'}(x')  g(x')\right)  \deepow{n} x'  \,,$$
where the integral on the boundaries is zero, since the derivatives of the probability measures are also suppressed at infinity. It is possible to do an external differentiation now, calculating the divergence of $E_\mu(x)$, and using eq.~\eqref{eq:LaplaceGreen}:

$$ \partial_\mu E_\mu(x) = \int \limits_{\mathbb{R}^n} \underbrace {-\frac{1}{S_n}\partial_{\mu} \partial_{\mu} \frac{1}{|x-x'|^{n-2}}}_{\Delta G(x,x') = \delta(x-x')} \cdot \partial_{\mu'} \left(\phi_{\mu'}(x')  g(x')\right)\cdot   \deepow{n} x' =  \partial_{\mu} \left(\phi_{\mu}(x)  g(x)\right) \,.$$

For an arbitrary $\phi_\mu$, $E_\mu$ and $g\phi_\mu$  can only differ by a divergence-free vector field. In order to show $|E| \approx g$, a certain measure is needed for the difference. With the square of the difference it becomes:

\begin{equation}\label{eq:DensityError}
\int \limits_{\mathbb{R}^n}\left ( g(x)\phi_\mu(x) - E_\mu (x) \right)^2  \deepow{n}x \geq 0    \,.
\end{equation}
Expanding the square, it will have three terms:

$$ \int \limits_{\mathbb{R}^n}\deepow{n}x   \underbrace{g^2}_{\text{invariant}} - 
	2 \underbrace{g\phi_\mu E_\mu}_{\text {dipole energy}} + \underbrace{E^2}_{\text{field energy}} \,.$$ 
The field energy can be expressed in terms of the dipole energy. Explicitly writing it out:

$$ \int \limits_{\mathbb{R}^n}\deepow{n}x E^2 =  \frac{1}{S_n^2}\int \limits_{\mathbb{R}^{2n}} \deepow{n}x \deepow{n}y  \left[ g^{x}\phi_{\mu}^x g^y\phi_{\nu}^y\right] \underbrace{\int \limits_{\mathbb{R}^{n}} \deepow{n}z \partial_{x_\mu} \partial_{z_\eta} \frac{1}{|x-z|^{n-2}} \partial_{y_\nu} \partial_{z_\eta} \frac{1}{|y-z|^{n-2}}}_{I_{\text{int}}}  \,.$$
Performing integration by parts on the $I_{\text{int}}$ term, $\Delta \frac{1}{|x-z|^{n-2}}$ appears. Due to eq.~\eqref{eq:LaplaceGreen} this can be substituted with a delta function. The remaining part has the derivative of the delta function, which can be easily evaluated:

$$ I_{\text{int.}} = - S_n \int \limits_{\mathbb{R}^{n}} \deepow{n}z \partial_{x_\mu} \delta(x-z) \partial_{y_\nu}  \frac{1}{|y-z|^{n-2}} =  S_n\partial_{y_\nu} \partial_{x_\mu} \frac{1}{|x-y|^{n-2}} \,.$$
As a result the field energy equals the dipole energy

$$\int \limits_{\mathbb{R}^n}\deepow{n}x E^2 =  \int \limits_{\mathbb{R}^n}\deepow{n}x g\phi_\mu E_\mu \text{  ,}$$
so the eq.~\eqref{eq:DensityError} can be modified accordingly:

$$ \int \limits_{\mathbb{R}^n}\left ( g(x)\phi_\mu(x) - E_\mu (x) \right)^2  \deepow{n}x = 
 \int \limits_{\mathbb{R}^n}\deepow{n}x g^2 - g\phi_\mu E_\mu
 \geq 0 \,.$$

The minimum is not necessarily zero, but it can be reached by finding the energy minimum of the dipole system. 

\subsection{Existence of the optimum}
\label{sec:ExistenceOfSolution}
It was assumed for eq.~\eqref{eq:identityfordensity}, that an identity operator exists for a given $g(x)$ probability density with a special $\phi_\mu(x)$ configuration:

$$ \partial_\mu c(x) = g(x)\phi_\mu(x) \,.$$
In other words, such a $c(x)$ has to exist which has a gradient with specified absolute value $|\partial_\mu c(x)| = g(x)$. This only exists if $g(x)\phi_\mu(x)$ can be made curl free, and all possible loop integrals should result in zero:

$$ \oint \limits_L g(x)\phi_\mu(x) \dee l_\mu = 0 \,.$$
As an $n$-dimensional cube has $n(n-1)$ faces, from which $n(n-1)/2$ share the same corner, $n(n-1)/2$ linearly independent infinitesimal loops can be imagined. So the curl in $n$-dimensions may have that many independent variables, which has to be zeroed out by a $\phi_\nu$ vector field with only $n-1$ independent variables. That seems to be non-trivial, but for differentiable $g(x)$, it turns out it can be done easily. The curl in question is:

\begin{equation}\label{eq:NDimCurlFree}
 \left[ \text{curl }g(x)\phi_\mu(x)\right]_{\nu\eta} = \partial_{(\nu)} (g\phi_\mu) e^{(\nu)}_\mu - \partial_{(\eta)} (g\phi_\mu) e^{(\eta)}_\mu \overset{?}{=} 0  \,,
\end{equation}
where $e^{(\nu)}_\mu$ denotes the unit basis vector in the $\nu$ direction, and the bracketed indices are \emph{not} summed up. In the basis $e^{(\nu)}_\mu = \delta_{\nu\mu}$, the curl is simplified:

$$  \left[ \text{curl }f(x)\phi_\mu(x)\right]_{\nu\eta} = \partial_{(\nu)} (g\phi_{(\nu)}) - \partial_{(\eta)} (g\phi_{(\eta)})  \overset{?}{=} 0 \,.$$
As only certain derivatives of $\phi_\mu$ appear, one possibility to anull the curl is to set every subcomponent to zero:

$$  \partial_{(\nu)} (g\phi_{(\nu)})  = 0$$
$$ \frac{\partial_{(\nu)}\phi_{(\nu)}}{\phi_{(\nu)}}= - \frac{\partial_{(\nu)} g} {g} \,.$$
In this differential equation only the relative change of $\phi_\nu$ appears, so setting $\phi_\mu \phi_\mu = 1$ at every $x$ is still possible. As a consequence, the $\phi_\mu$ vector field exists for any differentiable $g$, making $g\phi_\mu$ curl-free, and a $c(x)$ function exists with the property $\partial_\mu c(x) = g(x) \phi_\mu(x)$, making eq.~\eqref{eq:identityfordensity} an identity. 

%
%

\section{Density estimation from a finite sample}
\label{sec:DensityEstimation}
As long as a suitable $\phi_\mu(x)$ field is found that satisfies eq.~\eqref{eq:StoppingCriterium}, it can be used to estimate any differentiable probability density function from a sample $\{x_i\}$:
\begin{equation} \label{eq:SampledIntegral}
 g(x) = \lim_{N \rightarrow \infty} \left| \frac{1}{NS_n}\sum_{i=1}^N \partial_\mu \partial_\nu \frac{1}{|x-x_i|^{n-2}} \cdot \phi_\nu(x_i) \right|
\,.
 \end{equation}
However, since the Green's function has divergences, a finite sample may produce unnecessarily large fluctuations. Fortunately, volumes around most of the divergences can be neglected, as their integral is zero. A homogeneous $g(x)=g$, homotropic $\phi(x)_\mu = \phi_\mu$ sphere surface around a point $x$ gives no contribution:

\begin{align*}
 \int \limits_{S_n} \partial_\mu \partial_{\mu'} \frac{1}{|x-x'|^{n-2}} g \phi_{\mu'} \deepow{n} x' & = g\phi_{\mu'} \partial_\mu \int  \limits_{S_n} \partial_{\mu'}\frac{1}{|x-x'|^{n-2}} \deepow{n} x' \\
 	& = g\phi_{\nu} \partial_\mu \underbrace {\int  \limits_{S_n} \underbrace{- \frac{1}{|x-x'|^{n-1}}}_{const.} \hat{r}_{\mu'} \dee S_n}_{\int \limits_{S_n} \hat{r}_\mu' \dee S_n = 0 } \dee r'  = 0  \,.\\
\end{align*}
As a result, a small sphere of radius $\Delta R$ around the point of interest $x$ can be excluded from the summation in eq.~\eqref{sec:DensityEstimation}, 
since the error will go as $\partial_\mu (g(x)\phi_\nu(x)) O(\Delta R)$ for any differentiable $g(x)$ and $\phi(x)$. The only exception where the error is not 
$O(\Delta R)$ is for $\delta$ functions, so these points have to be treated independently. 

In order to have control on the fluctuations, one has to choose the $\Delta R$ radius of the exclusion volume in such a way that a $\delta r$ thick shell around 
this sphere contains enough sample points to estimate the integral belonging to that area. In this case the points in this layer will have a dominant contribution 
to the sum between $\{ \frac{1}{\Delta R^n}, \frac{1}{(\Delta R + \delta r)^n}\}$, since points at larger distances will have decreasing values. To have the 
contribution in the shell to be of the same order of magnitude, $\delta r$ has to be around $\Delta R/n$. This approximation is derived from the behaviour of the 
$(1+{}^1/_n)^{-n}$ function, which becomes nearly flat in $n$, with a value around ${}^1/_2$. From this it follows that if one needs $N_\text{large}$ number of 
points in this stable layer, approximately $n\cdot N_\text{large}$ points have to be discriminated, which eventually determines $\Delta R$. In practice one may 
choose a different $N_\text{large}$ for the iterative search of the $\phi_\nu(x_i)$ parameters and for the evaluation of the density estimation. 

\subsection{Finding the parameters}
\label{sec:ParameterSearch}

With a suitable $N_\text{discr}$, the set of $\{\phi_{\mu i}=\phi_{\mu}(x_i)\}$ that can be used for density estimations can be found with iterative search. The stopping criterium can be formulated with discretising eq.~\eqref{eq:StoppingCriterium}:

\begin{equation}\label{eq:SampleStopping}
 \phi_{\mu i} \parallel E_\mu(x_i) = \frac{1}{NS_n}\sum_{\substack{j \in \{j : |x_i - x_j| > R_\text{discr}^i\} \\  \left| \{ k: | x_i - x_k| < R_\text{discr}^i \} \right | = N_\text{discr}} }   \partial_\mu \partial_{\mu'} \frac{1}{|x_i-x_j'| }\phi_{\mu' j}  \,.
 \end{equation}

As this requirement is similar to looking for the ground state of the dipole system, a way to find it is the gradient descent on the energy function. With $E_{i\mu} = E_\mu(x_i)$ this direction is:

\begin{equation} \label{eq:GradientSteepestDescent}
\dee \phi_{i\eta} = (1-\phi_{i\eta} \phi_{i\mu} )E_{i\mu}  \,,
\end{equation}
which is a rotation with the $E_\mu$ field in the directions perpendicular to $\phi_\mu$. A gradient descent method typically contains a line search part, but it is not necessary here. It is enough to set a limit for the angular change of the $\phi_{i\mu}$, an order or magnitude smaller than the maximal possible, $|\dee \phi_\text{max} | = 0.1 < \pi $, seems to be a good choice, which ensures that the $E_\mu$ field won't change significantly during the iteration, keeping it a valid gradient. Further speed increase can be achieved, while taking care not to overshoot the angle between $E_{i\mu}$ and $\phi_{i\mu}$, since this is what minimises the energy. 
However, there might be several local energy minima, but the identity in eq.~\eqref{eq:identityfordensity} is only true if $g\phi$ is curl free. Following from the inequality in eq.~\eqref{eq:DensityError} it means that the non-global minima solutions are not satisfying the identity and they are not curl free. These solutions will hold a bias in the density estimations, which can be relatively big in the low density regions. This is because the non-optimal minima have a remnant polarisation, for example the destructive interferences do not fully cancel at zero density regions. 

\begin{figure}[h]
\begin{minipage}{18pc}
\includegraphics[width=18pc]{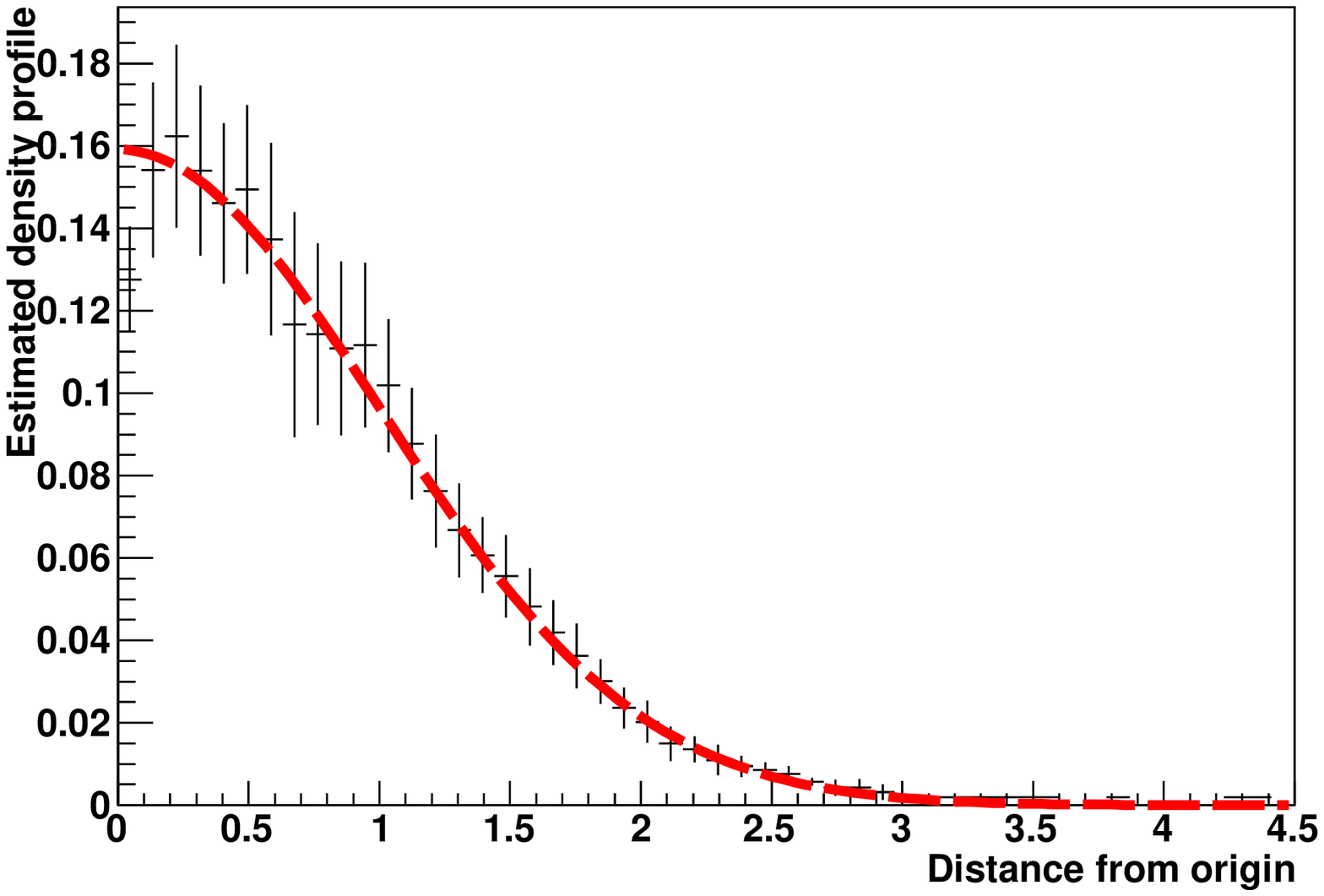}
\caption{\label{fig:GaussProfile}Radial density estimate profile of a Gaussian distribution placed at the origin. The crosses show the spread of the Green's function density estimate compared to the true value shown as the dashed red curve.}
\end{minipage}\hspace{2pc}%
\begin{minipage}{18pc}
\includegraphics[width=18pc]{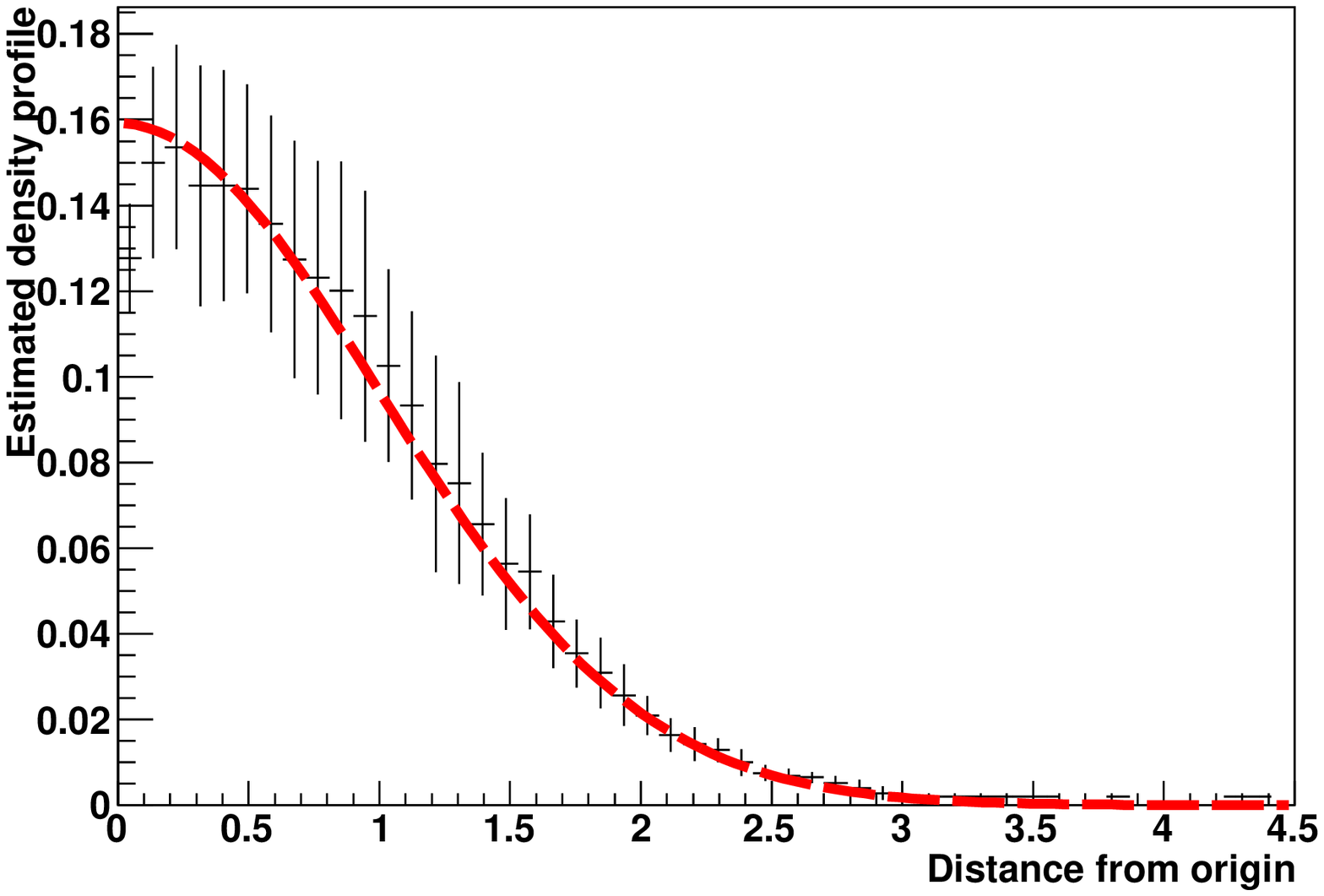}
\caption{\label{fig:GaussProfileNK}Profile plot for the density estimate made for the same Gaussian sample as in fig.~\ref{fig:GaussProfile} made with the nearest $k$-neighbour method with a flat kernel.\\}
\end{minipage} 
\end{figure}

As can be seen on fig.~\ref{fig:GaussProfile}, the algorithm actually performs as a density estimator. A sample with a size of two thousand points from a two dimensional Gaussian distribution was fed into the parameter search, and the density estimation was evaluated in the described manner, at the position of sample points. The figure shows the average of the estimated densities for a given radius, where the theoretical value is constant, because of the rotational symmetry of the chosen Gaussian distribution. The spread was calculated from the variance of the estimations on the same radius. The uncertainty can be directly compared with the simplest, kernel-less nearest $k$-neighbour method~\cite{Bishop:2006:PRM:1162264} in fig.~\ref{fig:GaussProfileNK}. It can be seen, that the uncertainty for the discussed density estimator in fig.~\ref{fig:GaussProfile} is approximately half of that from the nearest $k$-neighbour method with flat kernel in fig.~\ref{fig:GaussProfileNK}. The resolution were chosen to be the same for the two figures, given that the exclusion radii, defined in Section~\ref{sec:DensityEstimation}, were the same as the radii that was used for volume calculation in the nearest $k$-neighbour method.

\section{Classification and overtraining}

Although binary classification can be done with binary regression models, their results typically depend only on the fraction of the signal $s(x)$ and background density $b(x)$, which is optimal due to the Neyman-Pearson Lemma~\cite{Neyman01011933}.
This is also true for an optimally trained neural network with a quadratic loss function, where the response is usually the following~\cite{Bishop:2006:PRM:1162264}\cite{2004physics...2093F}:

$$ r_\text{NN} = \frac{s(x)}{s(x) + b(x)}\,.$$

In case the network is not optimally trained, the error of the response $r_\text{NN}$ can be related to the uncertainty of the $s(x)/b(x)$ fraction. The same fraction can be calculated from a density estimator, giving a likelihood estimation. The uncertainty in the latter case is much better under control, since the calculation of the signal and background densities are made independently. For the method described in this article, a density estimation for a single point comes from the superposition of the kernels of the most dominant $N_\text{large}$ sample points, hence by construction the likelihood estimation can not come from a single point, as happens during overtraining in regressions. 

\begin{figure}[h]
\begin{minipage}{18pc}
\includegraphics[width=18pc]{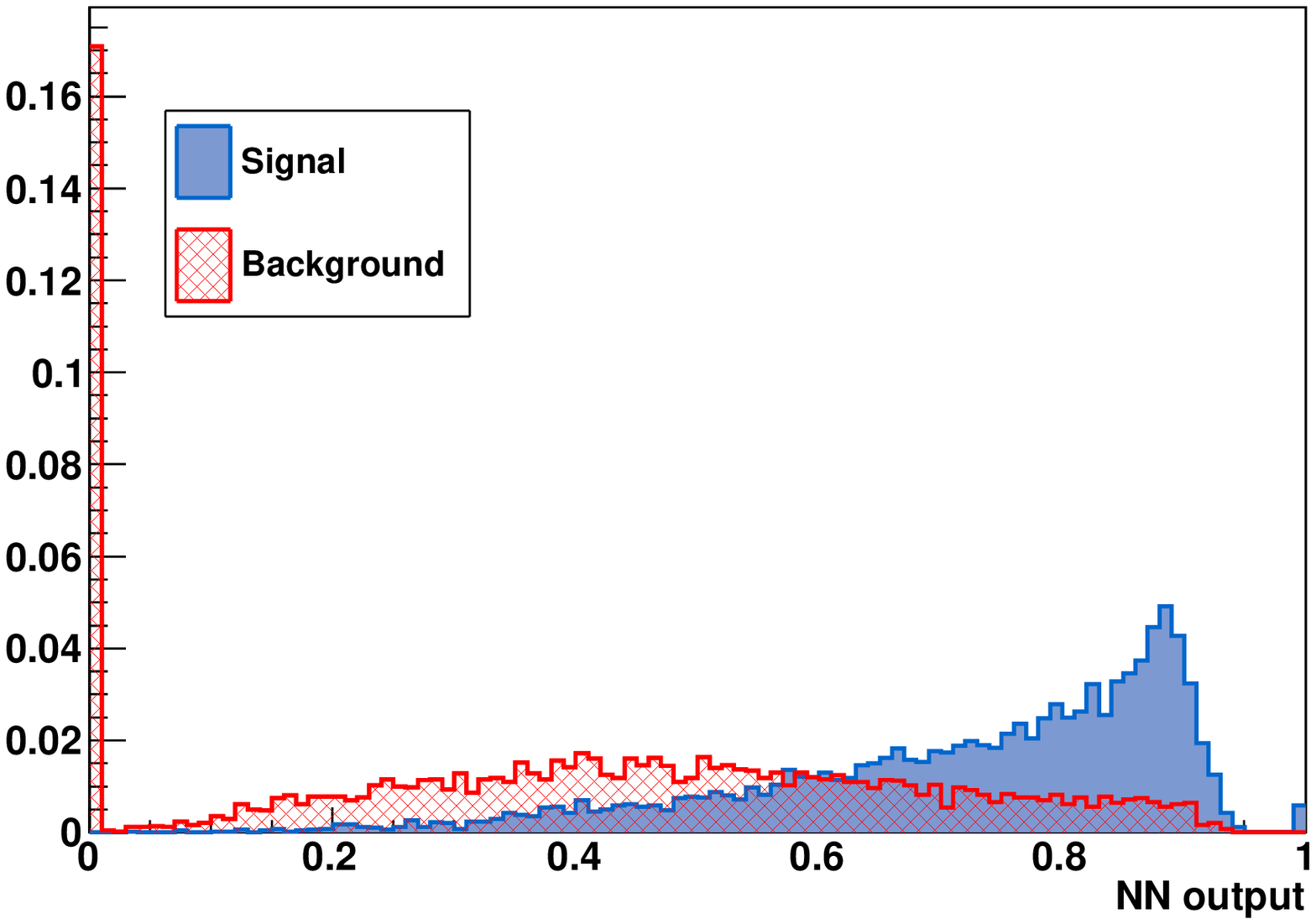}
\caption{\label{fig:TMVAself}Distribution of the response function on the $10^4$ training sample.}
\end{minipage}\hspace{2pc}%
\begin{minipage}{18pc}
\includegraphics[width=18pc]{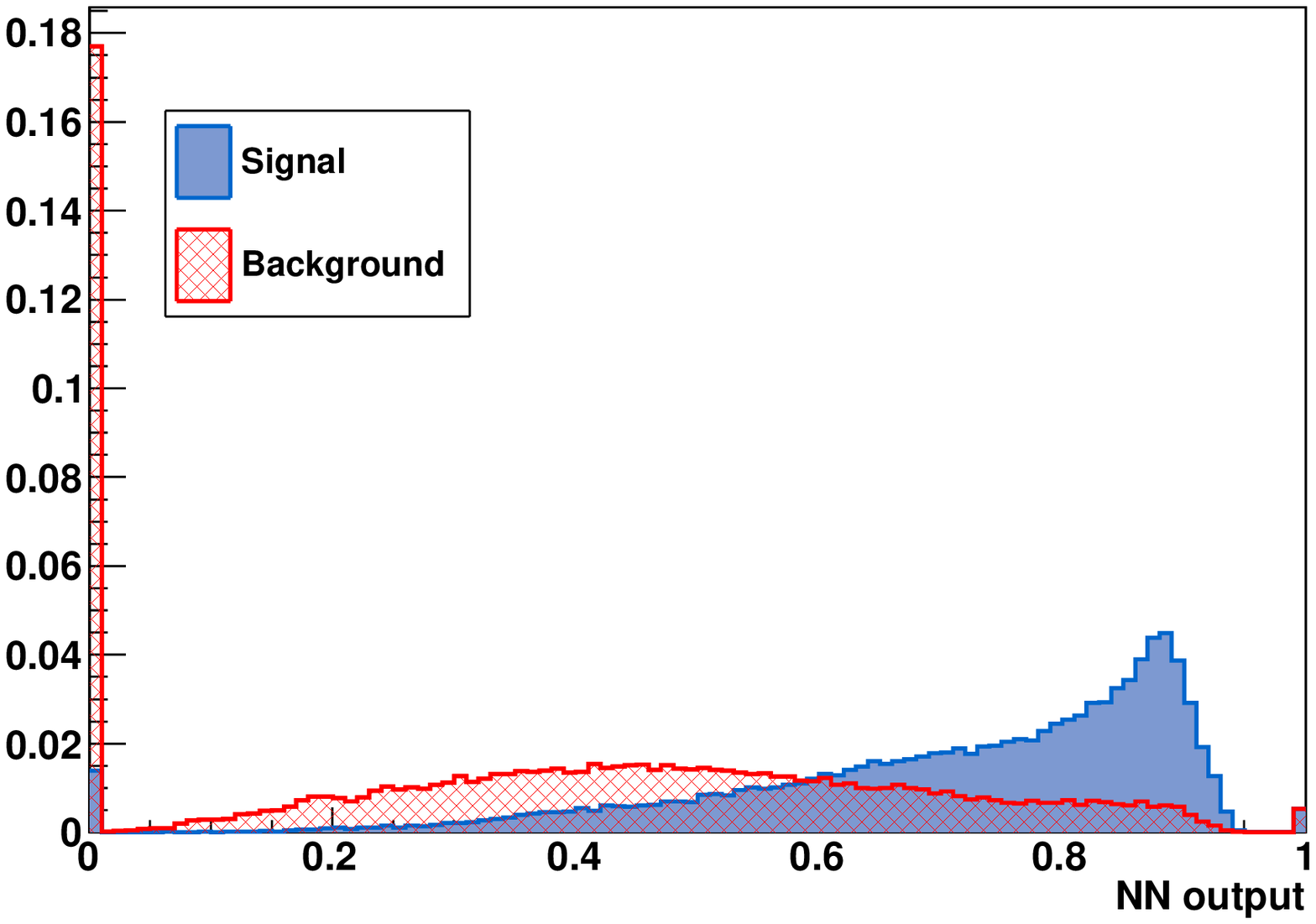}
\caption{\label{fig:TMVAtest}Distribution of the response on an independent $10^5$ test sample.}
\end{minipage} 
\end{figure}

Figure~\ref{fig:TMVAself} and fig.~\ref{fig:TMVAtest} show the response distribution for a network trained on a signal consisting of twelve non-overlapping Gaussians and a flat background~\cite{privateEckhardVT:2011j}. Unlike in many regression models, here there is no need for an independent test sample for an early stopping criterium. Figure~\ref{fig:TMVAself} is evaluated on the full training sample, while fig.~\ref{fig:TMVAtest} was made on an independent sample. The shape of the two responses are essentially same, the slight difference is at the part where either the signal or the background distribution density is very low, since these areas have the largest uncertainty in density estimators.

\section{Conclusions}

The article presents a novel class of density estimators, based on Green's function identities. 
The investigated one is derived from the Green's function of the Laplace operator which provides meaningful results for 
once differentiable probability density functions. It is possible to construct density estimators from other linear operators, but they are expected to require either higher order differentiability or other functional constraints on the investigated probability density. As it was shown, such a density estimator can be used for binary classification. In such case, due to the handling of the fluctuations, overtraining is not so prominent as it is in binary regression models. Therefore there is less need for independent test samples.

\section*{Acknowledgements}

We would like to thank Eckhard von T\"orne for giving encouragement, for pointing out the important topics needed to be touched and for the sample he prepared and gave us to use for testing the classification power of the algorithm.

\section*{References}
\bibliographystyle{BibTex/iopart-num/iopart-num}
\bibliography{BibTex/references.bib}

\end{document}